%% file: 00main.tex
\documentclass{article}
\usepackage[preprint]{corl_2022}

\usepackage{booktabs}
\usepackage{multirow}
\usepackage{subcaption}

\newcommand{\beginsupplement}{
    \setcounter{table}{0}
    \renewcommand{\thetable}{\Alph{table}}
    \setcounter{figure}{0}
    \renewcommand{\thefigure}{\Alph{figure}}
    \setcounter{section}{0}
    \renewcommand{\thesection}{\Alph{section}}
}

\usepackage[colorinlistoftodos,prependcaption,textsize=tiny]{todonotes}

\begin{document}
\title{SORNet: Spatial Object-Centric Representations for Sequential Manipulation}
\author{
    Wentao Yuan \\ University of Washington \And
    Chris Paxton \\ NVIDIA \And
    Karthik Desingh \\ University of Washington \And
    Dieter Fox \\ University of Washington, NVIDIA \AND \url{https://sites.google.com/view/sornet-extended}
}
\maketitle

\input{01abstract}

\input{02introduction}

\input{03background}

\input{04methods}

\input{05dataset}

\input{06results}

\input{07casestudy}

\input{09conclusion}

\bibliography{references}

\clearpage
\input{10supplement}

\end{document}

%% file: 01abstract.tex
\begin{abstract}
Sequential manipulation tasks require a robot to perceive the state of an environment and plan a sequence of actions leading to a desired goal state.
In such tasks, the ability to reason about spatial relations among object entities from raw sensor inputs is crucial in order to determine when a task has been completed and which actions can be executed.
% Prior works relying on explicit state estimation or end-to-end learning struggle with novel objects or new tasks. 
In this work, we propose \textbf{SORNet} (\textbf{S}patial \textbf{O}bject-Centric \textbf{R}epresentation \textbf{Net}work), a framework for learning object-centric representations from RGB images conditioned on a set of object queries, represented as image patches called canonical object views. With only a single canonical view per object and no annotation, SORNet generalizes \emph{zero-shot} to object entities whose shape and texture are both \emph{unseen} during training. We evaluate SORNet on various spatial reasoning tasks such as spatial relation classification and relative direction regression in complex tabletop manipulation scenarios and show that SORNet significantly outperforms baselines including state-of-the-art representation learning techniques.
We also demonstrate the application of the representation learned by SORNet on visual-servoing and task planning for sequential manipulation on a real robot. Our code and data are available publicly at \url{https://github.com/wentaoyuan/sornet}.
\end{abstract}
\keywords{Object-centric Representation, Spatial Reasoning, Manipulation}

\begin{figure}[h]
    \centering
    \includegraphics[width=\linewidth]{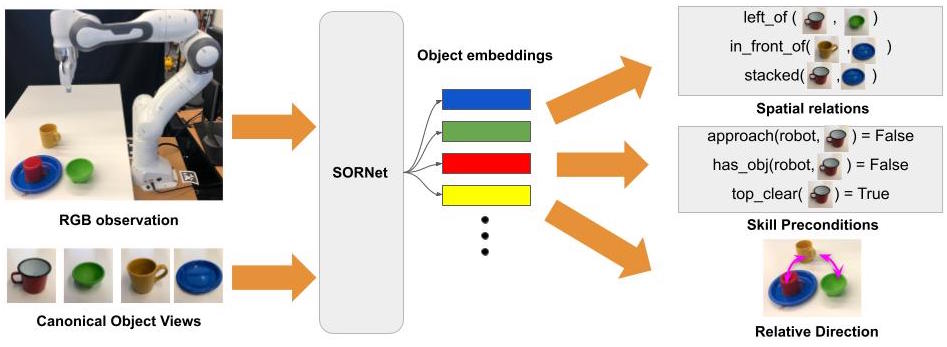}
    \caption{
        We propose \textbf{SORNet} (\textbf{S}patial \textbf{O}bject-Centric \textbf{R}epresentation \textbf{Net}work), a method which learns object embeddings from RGB observation given a set of object queries called canonical object views. SORNet adapts to novel objects without the need for annotating new data and finetuning. SORNet embeddings can be used to solve a variety of spatial reasoning tasks such as classifying spatial relations and regressing relative directions between objects.
    }
    \label{fig:cover}
\end{figure}

%% file: 02introduction.tex
\section{Introduction}
Robots must be able to understand the notion of objects and the relationship between them in order to complete many complex multi-step tasks. Take, for example, a task where the robot needs to destruct a tower of blocks and rebuild it in a different order. The robot has to be able to tell whether each block is accessible, determine how to make each block is accessible, and understand the consequences of actions as each block is moved. 

To solve those long-horizon sequential manipulation tasks, it is common to use a state estimator followed by a task and motion planner~\cite{fikes1971strips, fox2003pddl2, rovida2017extended, paxton2019representing, sui2017goal, balakirsky2012industrial}. These model-based systems are powerful at reasoning and apply to many different tasks with various goal conditions.
However, they are limited by the ability of the state estimator, which outputs explicit object state such as 6D poses that are difficult to estimate precisely from raw data and not optimized for downstream tasks. Although a variety of powerful approaches exist for explicitly estimating the state of objects in the world~\cite{xiang2017posecnn,sundermeyer2018implicit,li2018deepim,deng2021poserbpf},
it is challenging to generalize these approaches to an arbitrary collection of objects. Further, manipulation scenes often include contact and occlusion in which state estimation methods tend to fail~\cite{xie2020best,xiang2020learning}.

Fortunately, knowing explicit states such as exact poses of objects is not necessary for manipulation. There is another line of works showing that it is possible to learn motor controls directly from raw sensor data, e.g. RGB images and joint encoder readings. Often known as the end-to-end approach, these methods leverage powerful neural network backbones that are able to extract low-level embedding vectors from high-dimensional images and directly optimize for downstream tasks. While there are recent methods in this category~\cite{levine2016end, ghadirzadeh2017deep, duan2017one, huang2019neural, paxton2019prospection, yu2018one} that can perform complex manipulation tasks, they are often not transferable to new tasks and goals, especially those involving long-horizon planning. We believe the key limitation is that these methods use scene-level representations (i.e. a single embedding vector per image) that do not support reasoning at an object level.

In this work, we take an important step towards an object-centric representation learning framework has explicit notion of objects but implicit, learnable embeddings that generalize to novel objects, can be optimized for downstream tasks, and support object-level reasoning for a variety of goal-conditioned tasks. Specifically, we propose a neural network backbone called \textbf{SORNet} (\textbf{S}patial \textbf{O}bject-Centric \textbf{R}epresentation \textbf{Net}work). SORNet extracts object embeddings from raw RGB observations conditioned on a set of object queries, which we call canonical object views. The design of SORNet, shown in Fig.~\ref{fig:cover}, allows it to generalize to scenes with novel objects without any change in its parameters. The object-centric embeddings produced by SORNet can be combined with readout networks to inform a task and motion planner with important spatial relations needed to figure out a plan for goal-directed sequential manipulation tasks, such as logical preconditions for primitive skills or continuous 3D directions between object centers.

To summarize, our contribution are:
(1) a method for extracting object-centric embeddings from RGB images that generalizes zero-shot to different number and type of objects;
(2) a framework for learning object embeddings that capture both logical and continuous spatial relations;
(3) a dataset containing RGBD sequences of various tabletop rearrangement manipulation tasks labeled with spatial predicates.

We empirically evaluate SORNet on classification of logical predicates and regression of relative 3D direction between entities in manipulation scenes, and show the advantage of SORNet's object-centric representation over the scene-level representation produced by state-of-the-art pretraining methods on these spatial reasoning tasks. In addition, we test SORNet on held-out objects that did not appear in training data and demonstrate that SORNet generalizes to common household objects from novel categories in an \textit{zero-shot} fashion. For additional results and visualizations, please visit the project website at \url{https://sites.google.com/view/sornet-extended} .

%% file: 03background.tex
\section{Related Work}

\paragraph{Model-based Sequential Manipulation}
Most classical works on goal-directed sequential manipulation use a pipeline approach. The first step is a model-based state estimator, which outputs explicit object state such as bounding boxes or 6D poses from sensor observations and object models (e.g. 3D meshes). The second step is a task and motion planner~\cite{fikes1971strips, fox2003pddl2, rovida2017extended, paxton2019representing, sui2017goal, balakirsky2012industrial} which reason on top of the explicit object states to produce plans and actions leading to the goal. The limitations of this type of approaches is that the state estimator is often restricted to a fixed of objects.

\paragraph{End-to-end (model-free) Manipulation}
The end-to-end approach learns motor controls directly from raw sensor observations~\cite{levine2016end, ghadirzadeh2017deep, duan2017one, huang2019neural, paxton2019prospection, yu2018one}, skipping the need for object state estimation. While these methods avoid the reliance on object models and explicit states, they do not have the notion of objects at all. As a result, they lack the ability to perform reasoning and are often limited to simple environments with one or two objects and short-horizon tasks.

\paragraph{Learning Spatial Relations} 
Learning spatial relations between object entities have been studied in the field of 3D vision and robotics. 
In 3D vision, methods such as \cite{rosman2011learning,fichtl2014learning,sharma2020relational} predict discrete or continuous pairwise object relations from 3D inputs such as point clouds or voxels, assuming complete observation of the scene and segmented objects with identities.
In contrast, our approach does not make any assumptions regarding the observability of the objects and does not require pre-processing of the sensor data such as segmentaiton. 
In robotics, the learning framework proposed by Kase et al.~\cite{kase2020transferable} is most related to our approach. Their method takes a sequence of sensor observations and classifies a set of pre-defined relational predicates which is then used by a symbolic planner to produce a plan composed by a sequence of primitive skills. Compared to our approach, theirs is limited by the number and type of objects it can be applied to. The work by Migimatsu et al.~\cite{migimatsu2021grounding} is also similar, which trains specifically to predict predicates but can use looser supervision.

\paragraph{Visual Reasoning}
Recently, several advancements have been made on visual reasoning benchmarks \cite{johnson2017clevr,CLEVRER2020ICLR,girdhar2020cater} using transformer networks \cite{vaswani2017attention}.
Toward solving spatio-temporal reasoning task from CLEVRER~\cite{CLEVRER2020ICLR} and CATER~\cite{girdhar2020cater}, Ding et al.~\cite{ding2020object} proposed an object-based attention mechanism and Zhou et al.~\cite{zhou2021hopper} proposed a multi-hop transformer model. Both works assume a segmentation model to produce object segments and perform language grounding to the segments in order to answer reasoning questions.
Our SORNet architecture is simpler and can solve spatial-reasoning tasks for unseen object instances without requiring a segmentation or object detection module. Furthermore, our work focuses on a relatively complex manipulation task domain involving manipulator in the observations. Although our current work focuses on spatial relations in a single RGB frame, the object-centric embeddings could potentially be used for solving temporal-reasoning tasks.

%% file: 04methods.tex
\begin{figure}[t]
  \centering
  \includegraphics[width=\linewidth]{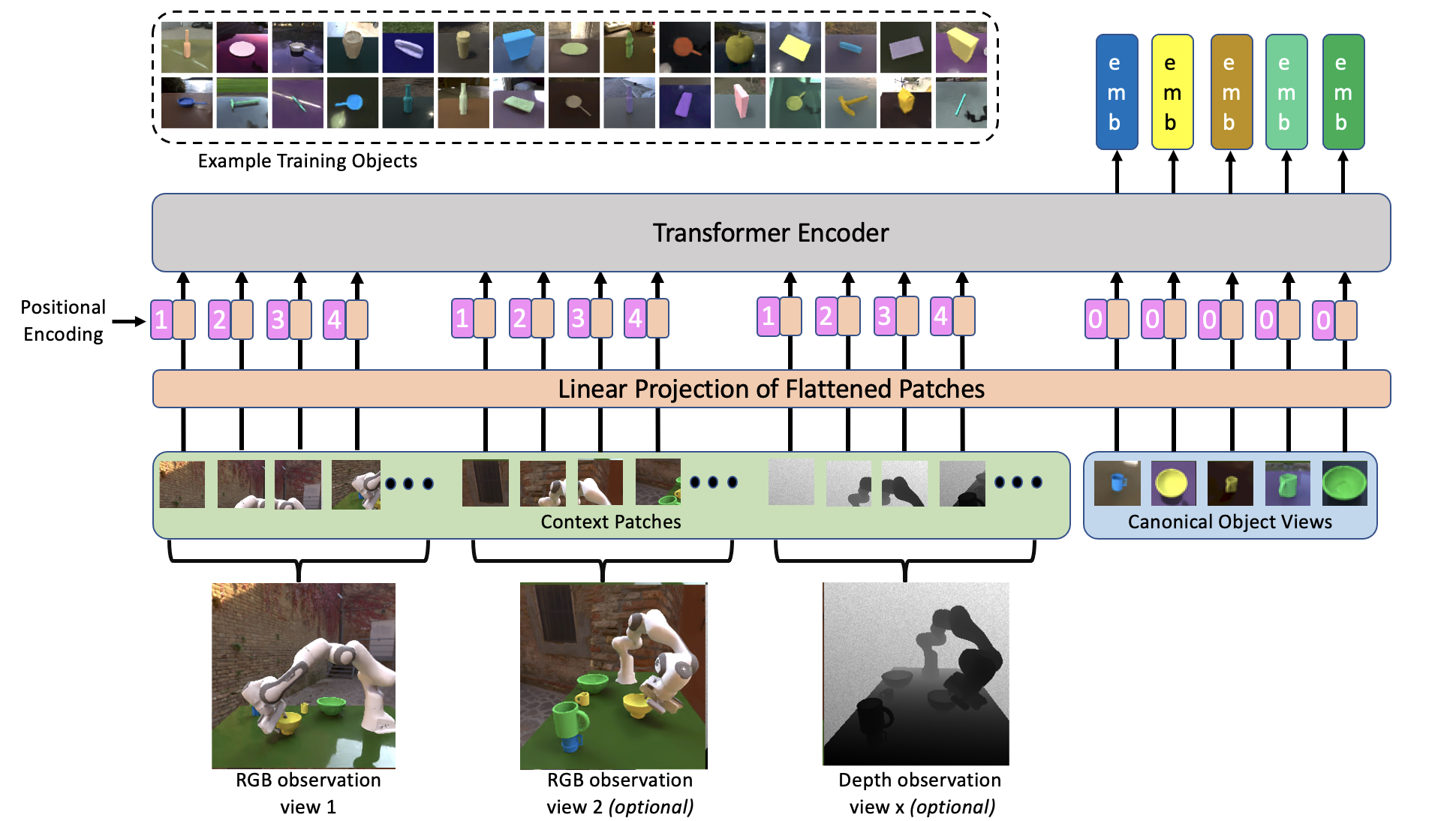}
  \caption{\textbf{SORNet} architecture. Input to the network is an RGB image and canonical views of the object queries. The RGB image is broken into context patches which have the same size as the canonical object views. The sequence of patches are flattened, position-encoded and passed through a multi-layer transformer to obtain a sequence of embedding vectors. The embeddings corresponding to the canonical object views are used for downstream tasks such as relation prediction. Additional views and modes of observations such as depth can be optionally added to the network. The top left inset shows examples of canonical object views used during training.}
  \label{fig:embedding_net}
\end{figure}

\begin{figure}
    \centering
    \includegraphics[width=0.49\linewidth]{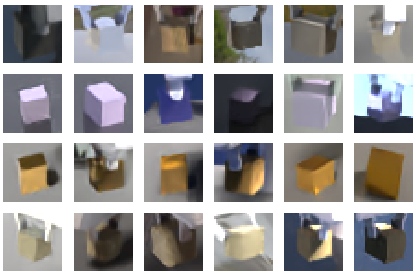}
    \includegraphics[width=0.49\linewidth]{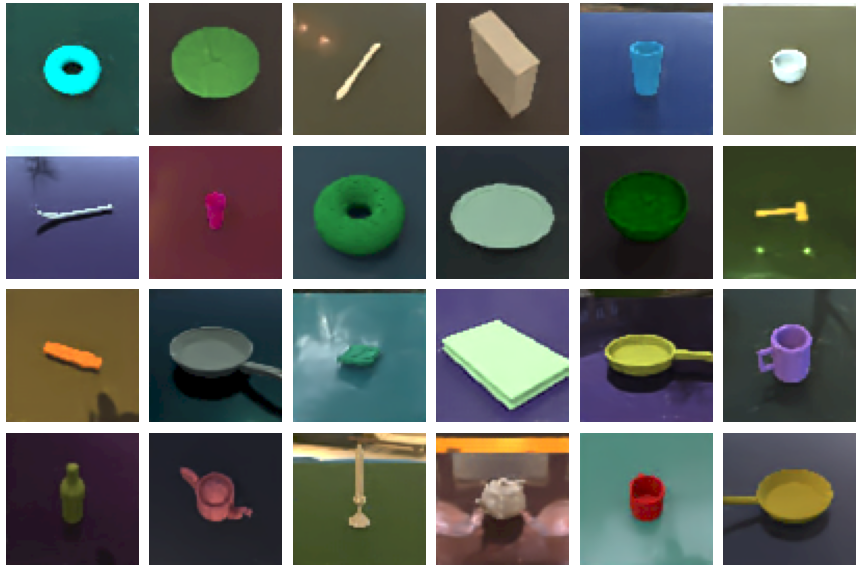}
    \caption{Examples of canonical object views in the Leonardo (left) and Kitchen (right) dataset. Lighting, texture and object poses vary across different views. Sometimes there is also occlusion by the robot.}
    \label{fig:obj_patches}
\end{figure}

\section{Methods}
\subsection{Object Embedding Network}
Our object embedding network (SORNet) (Fig.~\ref{fig:embedding_net}) takes an RGB image and an arbitrary number of object queries and outputs an embedding vector corresponding to each object query. The object queries are represented as image patches with a single object, which we call \textit{canonical object views}. Note that the canonical object views are \emph{not} crops from the input image, but arbitrary views of the objects of interest that may not match the objects' appearance in the scene. Our model learns to identify objects even under drastic change in lighting, pose and occlusion (see Fig.~\ref{fig:obj_patches} for examples of canonical object views used in our experiments). Intuitively, the canonical object views serve as queries where the RGB image serve as the context from which spatial relations are extracted. 

The architecture of the network is based on the Vision Transformer (ViT) \cite{dosovitskiy2020image}. The input image (or images) is broken into a list of fixed-sized patches, which we call \textit{context patches}. The context patches are concatenated with the canonical object views to form a patch sequence. Each patch is first flattened and then linearly projected into a token vector. We use add a list of learnable vectors with the same dimension as the token vectors as positional embeddings. The positional-embedded tokens are then passed through a transformer encoder, which includes multiple layers of multi-head self-attention \cite{vaswani2017attention}. The transformer encoder outputs a sequence of embedding vectors. We discard the embeddings from the context patches and keep those from the canonical object views as output object embeddings.

There are two implementation details we would like to mention. First, to increase spatial resolution, we find it beneficial to use smaller patch sizes in the transformer's input sequence, which can be smaller than the size of canonical object views. Consequently, there will be multiple tokens in the output sequence corresponding to the same object query. We simply concatenate the tokens coming from the same object as the representation for that object.

Second, the positional embedding is meant to inform the transformer the absolute position of the patch in the image. However, there is no inherent order among the set of object queries. In order to make the output object embeddings permutation equivariant, we force the set of position embedding vectors to be the same for each object (as suggested by the 0s in front of input object tokens in Fig.~\ref{fig:embedding_net}. In this way, we can pass in an arbitrary number of canonical object views in arbitrary order without changing model parameters during inference.

\subsection{Readout Networks} \label{sec:classifier}
\begin{figure}[t]
  \centering
  \includegraphics[width=\linewidth]{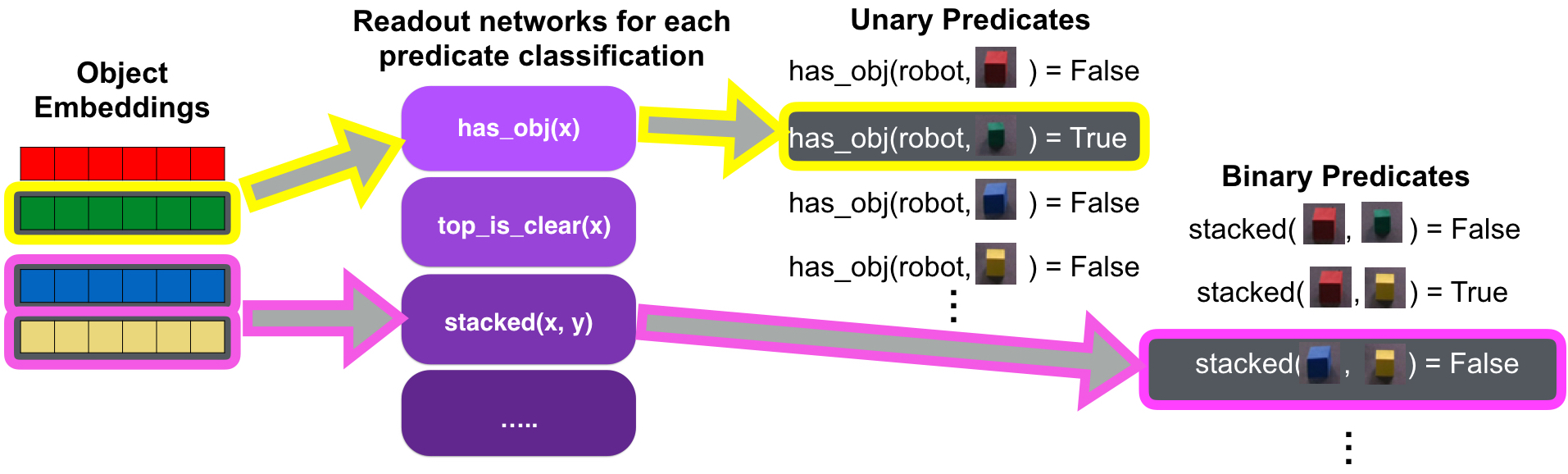}
  \caption{Architecture of the readout networks, which uses the object embeddings from \textbf{SORNet} to predict spatial relations, such as logical statements that can serve as skill preconditions or continuous 3D directions. The readout network is flexible to accommodate any number of input object embeddings without changing its parameters.}
  \label{fig:readout_net}
\end{figure}

The readout networks (Fig.~\ref{fig:readout_net}), as suggested by their names, are responsible for ``reading out" object relations using the embedding vectors produced by the embedding network. The relations can be logical statements, e.g., whether the blue block is stacked onto the green block. The relations can also be continuous quantities, e.g. which direction should the end effector move to reach the red block. The readout networks consist of a collection of 2-layer MLPs, one for each type of relations. Each readout network takes a number of object embeddings as inputs and outputs a quantity relevant to the input object arguments.

Here we focus on unary and binary relations, but our framework is extensible to relations involving more than two objects. Unary readout networks take a single object embedding as input and output a quantity involving a single object and optionally an environment element, which could be the robot or a region on the table. Taking the \texttt{top\_is\_clear} classifier for an example. If the input embedding is conditioned on the blue block, the network will output whether there is any object on top of the blue block. Binary readout networks take a list of binary object embeddings created by concatenating pairs of object embeddings and outputs quantities involving a pair of objects, e.g., whether the blue block is on top of the red block.

An important property of our design of readout networks is that the parameters of the readout network are independent to the number of object inputs. With $N$ input object embeddings, each unary readout network will output $N$ relations and each binary readout network will output $N(N-1)$ relations. The number of output relations dynamically changes with the number of inputs. In this way, at inference time, our overall model generalizes zero-shot to scenes where the number of objects is different from training.

%% file: 05dataset.tex
\section{Data Generation} \label{sec:data}
In order to force the network to learn spatial relations between the objects rather than rather than overfitting to irrelevant factors like color and texture, it is necessary to train it on a large-scale dataset with diverse object appearance and scene layouts. Existing robotics datasets are often limited by object variety (e.g. only has YCB objects \cite{calli2015ycb}) and scene complexity and existing vision datasets do not contain robot or object layouts that are interesting in manipulation (e.g. stacked objects). Thus, we create our own tabletop environment where a Franka Panda robot manipulates a set of randomly colored objects.

\begin{figure}
    \centering
    \includegraphics[width=\linewidth]{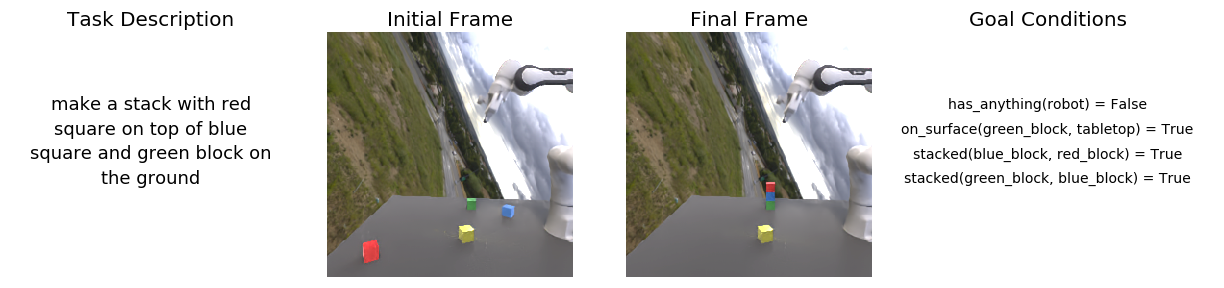}
    \includegraphics[width=\linewidth]{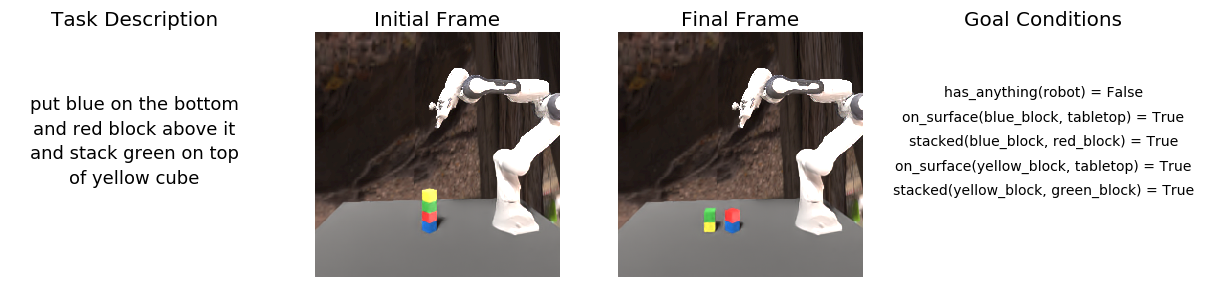}
    \includegraphics[width=\linewidth]{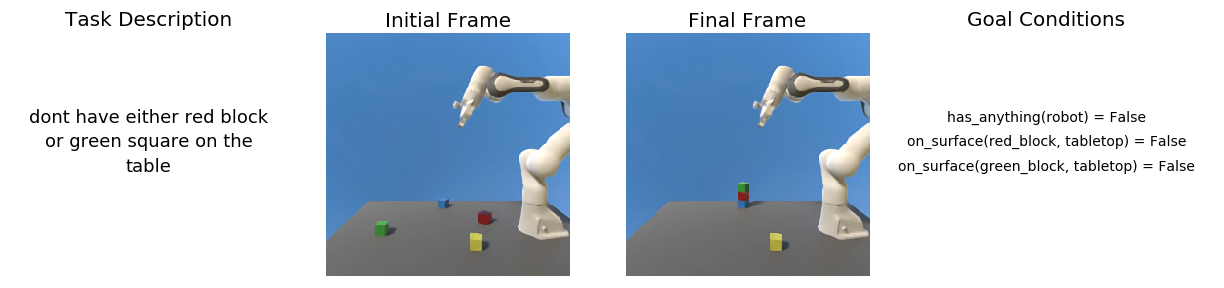}
    \includegraphics[width=\linewidth]{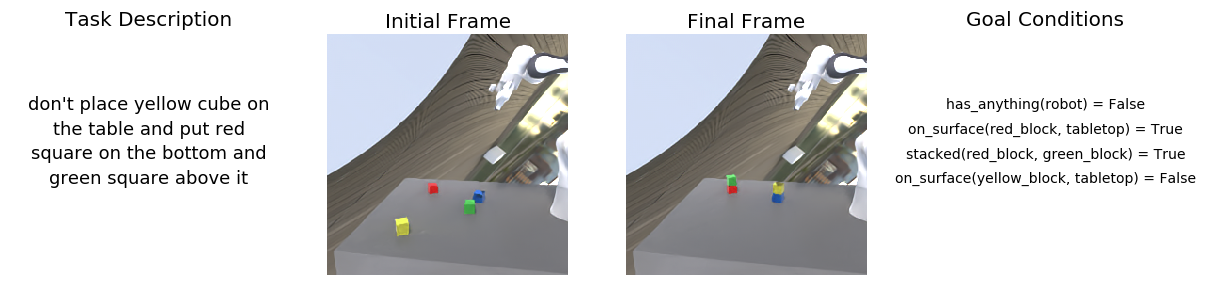}
    \includegraphics[width=\linewidth]{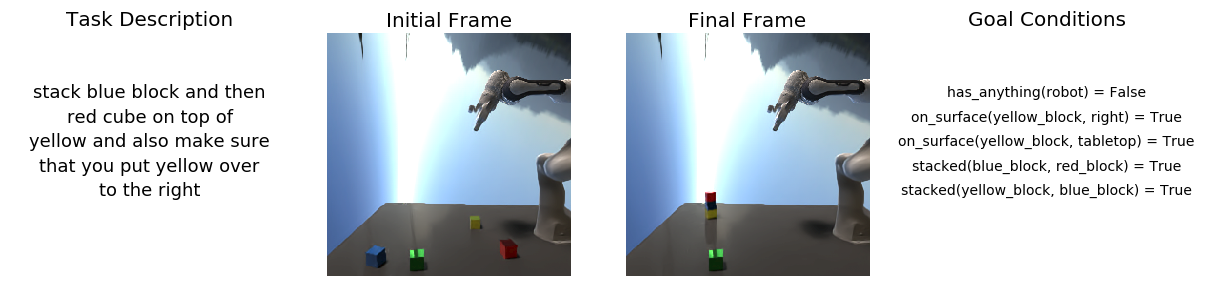}
    \caption{Visualization of some of the different tasks in the Leonardo test set. We used a wide variety of tasks to generate interesting data, taking the form of building one or more towers of varying heights, ensuring that certain objects were or were not on the table, and combinations of the above.}
    \label{fig:tasks}
\end{figure}

In our data, the robot is given a goal formulated as a list of logical predicates to be satisfied. A simple task and motion planner is used to find a plan~\cite{paxton2019representing} given ground truth object states in simulation. Fig.~\ref{fig:tasks} shows some example tasks in the data. 
Then, the robot executes the plan in simulation and we use NVISII~\cite{Morrical20visii} to render the RGB and depth images every time a predicate value is changed. This ensures that we capture the important moments when the robot transitions to a different primitive skill. Domain randomization including random lighting, background and perturbations to the camera position was applied while rendering to increase visual diversity. Ground-truth logical state is computed and recorded with the rendered frame.

We create two datasets, Leonardo and Kitchen. The Leonardo dataset has a single object shape (blocks) but more complex scene layouts (e.g. a tower of 4 stacked blocks). The Kitchen contains a large variety of object shapes (common household objects from the ACRONYM subset~\cite{eppner2021acronym} of ShapeNet~\cite{chang2015shapenet}). More details are elaborated below.

\begin{figure}
  \centering
  \includegraphics[width=\linewidth]{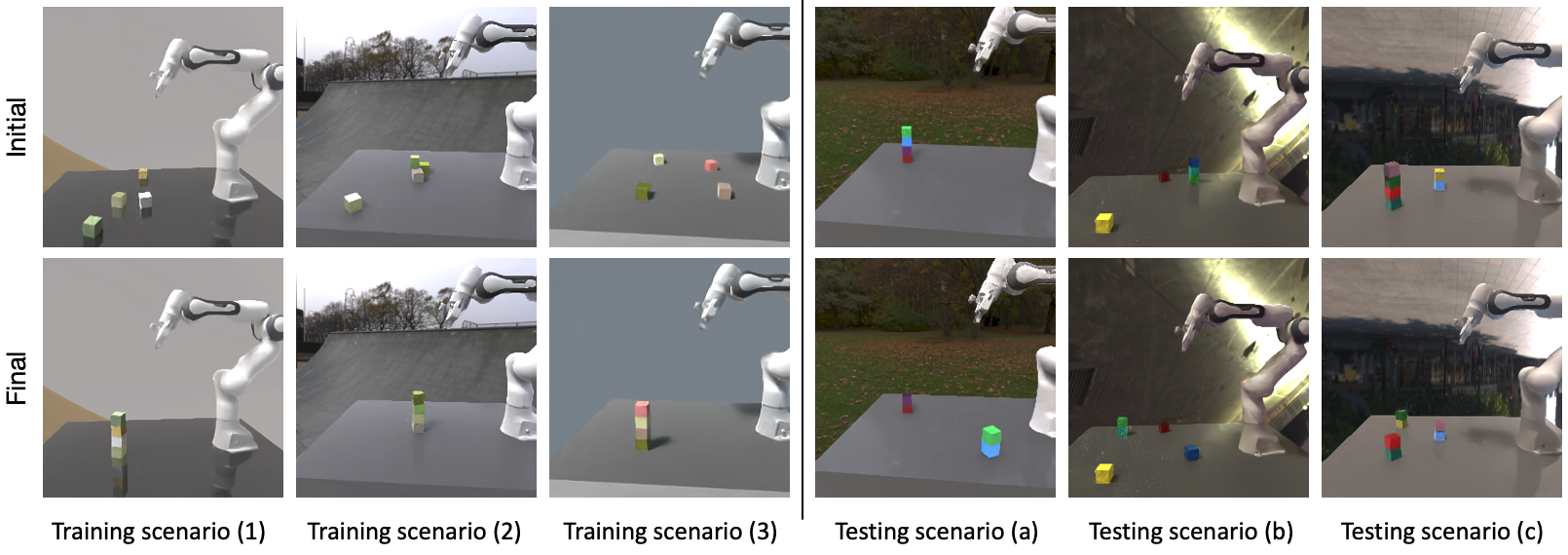}
  \caption{Sample scenes from training and testing scenarios in the Leonardo dataset. Top row shows the initial configuration of a sequence and the bottom row shows the goal configuration. The training scenarios contain 4 blocks with a single goal condition. The testing scenarios contain 4-7 blocks with heldout colors and various goal conditions involving multi-tower stacking.}
  \label{fig:leonardo_data}
\end{figure}

\begin{figure}
    \centering
    \includegraphics[width=\linewidth]{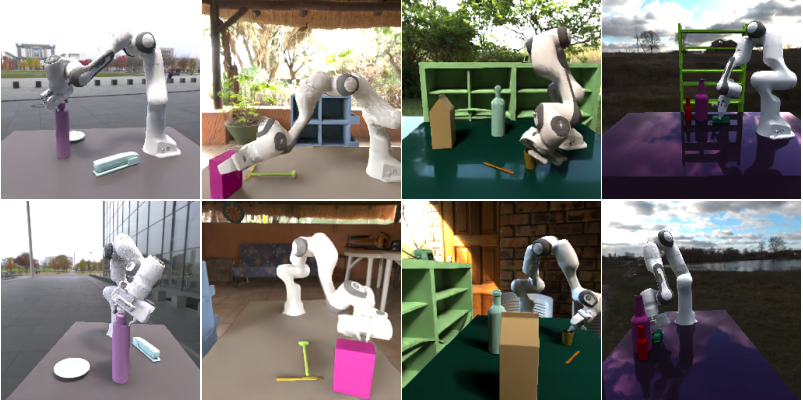}
    \caption{Sample scenes from the kitchen dataset. The top and bottom rows show two different views. SORNet can leverage additional views to improve performance, but does not require multiple views.}
    \label{fig:kitchen_data}
\end{figure}

\subsection{Leonardo Dataset}
\label{ssec:leonardo}

The first of our two datasets is the Leonardo blocks stacking dataset. The training data contains over 130K sequences of a single task - stacking 4 blocks in a tower. The block colors are randomly chosen from xkcd colors\footnote{\url{https://xkcd.com/color/rgb}}.
The testing data contains 1.6K sequences. Each sequence contains 4-7 blocks. The block colors are chosen from 7 colors that are not included in the training data: \textit{red, green, blue, yellow, aqua, pink, purple}. We also heldout all the colors whose name contains the test colors (e.g. \textit{light\_red}). This leaves 405 training objects in total.

The testing tasks consists of eight super-tasks which were all different from the single-tower-stacking task in the training data. These were: (1) building a single tower out of two blocks, (2) building two towers of two blocks, (3) making sure one specific block is not on the table, (4) making sure two blocks are not on the table, (5) building a tower of three blocks, (6) placing a block in a specific location, (7) making sure a block is not on the table and separately building a tower of two blocks, and (8) building a tower on a particular part of the table. Fig.~\ref{fig:leonardo_data} shows some examples of these tasks from the dataset.

\subsection{Kitchen Dataset}

The second of our two datasets is the Kitchen object rearrangement dataset. The objects in the dataset are taken from the set of Shapenet objects~\cite{chang2015shapenet} used in the ACRONYM grasp dataset~\cite{eppner2021acronym}, containing 330 object shapes from 33 categories that are likely to be found in a home, ranging from books to video game controllers. The Kitchen data use the same train/test color split as in Leonardo, but in addition to that, we also held out the mug and bowl classes to only appear in test data. This lets us show that SORNet generalize not only to unseen object instances, but to entirely unseen object categories. We also optionally add a shelf beside the table to increase environment diversity.

Instead of sampling a specific high-level goal as in Leonardo, we sampled high-level action traces of picking and placing random objects. Objects could be placed either in a table region, on a shelf (if it was available), or above another object. The training data contains over 24K sequences where the test data has 1.6K sequences. Each sequence has 3 to 7 objects appearing simultaneously. Fig.~\ref{fig:kitchen_data} shows some example frames.

%% file: 06results.tex
\section{Results}
In this section we discuss the comparisons between SORNet's object-centric embeddings and the scene-level embeddings learned by state-of-the-art representaiton learning techniques. We also demonstrate SORNet's unique ability to generalize to novel objects without any label.

\begin{figure}
    \centering
    \begin{subfigure}{\linewidth}
        \includegraphics[width=\linewidth]{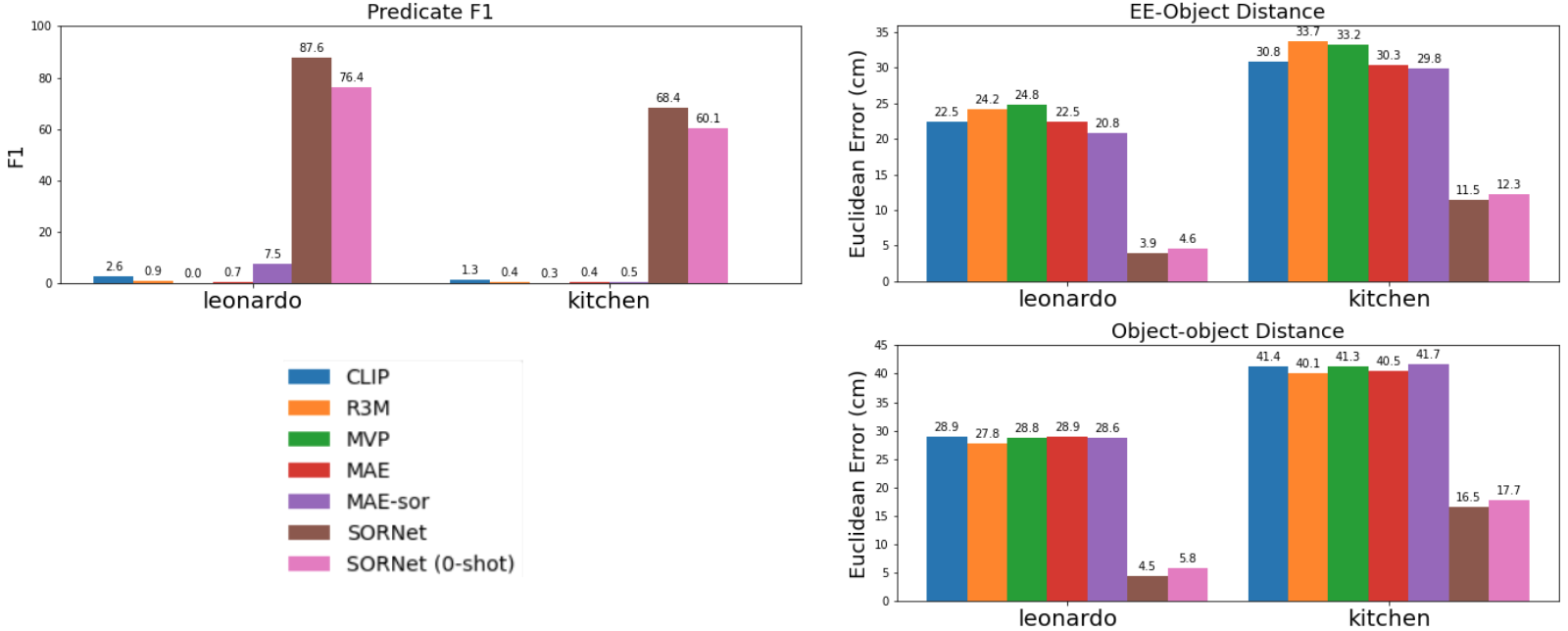}
        \caption{Frozen Embedding}
    \end{subfigure}
    \begin{subfigure}{\linewidth}
        \includegraphics[width=\linewidth]{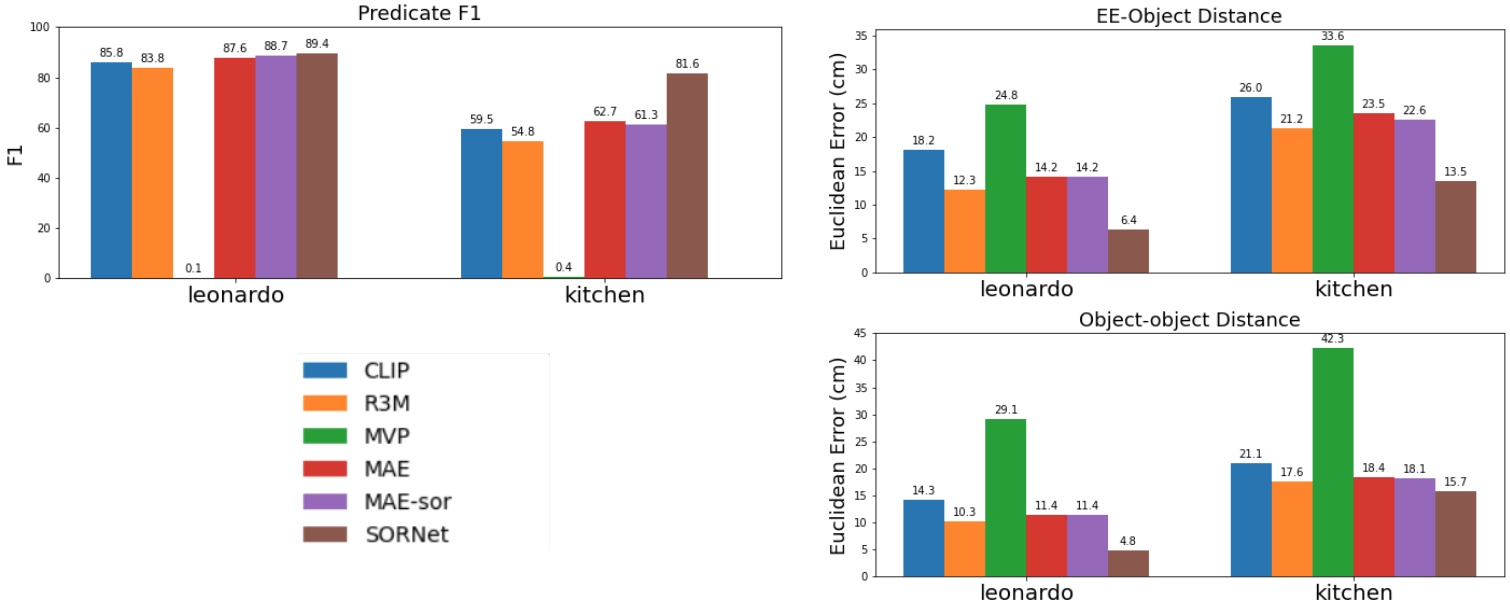}
        \caption{Finetuned Embedding}
    \end{subfigure}
    \caption{Predicate classification (left) and relative direction regression (right) results on Leonardo and Kitchen data. SORNet clearly outperforms other pre-training methods on reasoning about spatial relations in scenarios with complex object interactions. In addition, the 0-shot model that has not seen any labeled data for the test objects also performs reasonably well.}
    \label{fig:quant}
\end{figure}

\subsection{Spatial Relation Classification} \label{sec:predicate}
First, we evaluate SORNet on the task of predicting spatial relations among objects in manipulation scenes in the form of logical predicates, e.g. \texttt{left\_of(red\_mug, green\_bowl)}. The models are given an RGB image of the scene and are asked to predict a list of binary (True or False) predicates. Please refer to the supplementary for a complete list of predicates.

We benchmark our model against the following state-of-the-art pretraining methods.
\begin{itemize}
    \item \textbf{CLIP} \cite{radford2021learning} performs contrastive pretraining on a large-scale non-public image-caption dataset and has achieved state-of-the-art result on transferring to tasks like ImageNet classification.
    \item \textbf{R3M} \cite{nair2022r3m} pretrains a ResNet \cite{he2016deep} on egocentric videos from the Ego4D dataset \cite{grauman2022ego4d} using time contrastive loss and video-language alignment.
    \item \textbf{MVP} \cite{xiao2022masked} pretrains vision transformers \cite{dosovitskiy2020image} on egocentric videos from a combinary of datasets (e.g. Epic Kitchens \cite{damen2020epic}) using masked autoencoding (see next bullet).
    \item \textbf{MAE} \cite{he2021masked} pretrains visual transformers \cite{dosovitskiy2020image} on the ImageNet data using reconstruction from a masked input (referred to as masked autoencoding) as supervision and demonstrated state-of-the-art transfer results to image classification.
    \item \textbf{MAE-sor} is a variant of MAE that is trained on our data (Leonardo and Kitchen) using the masked autoencoding objective.
\end{itemize}

We design two evaluation protocols
\begin{itemize}
    \item \textbf{Frozen Embedding} In this case, we freeze the pretrained image embedding network and train readout networks (2-layer MLPs) only. This is to verify how much spatial information the pretrained embedding encodes.
    \item \textbf{Finetuned Embedding} In this case, we train the embedding network together with the readout networks. This is to see if the network has the ability to learn spatial information if provided with the right supervision.
\end{itemize}
To ensure a fair comparison, we trained all baselines and SORNet on 10K sequences where the objects are sampled from a fixed repository and tested on 1.6K sequences. The evaluation is done on both Leonardo (blocks only) and Kitchen data (common household objects). In Leonardo data, the repository has 7 objects and each scene has 4-7 objects. In Kitchen data, the repository has 14 objects and each scene has 3-8 objects. Both the baselines and SORNet are given ground truth object identity. The baselines always predict predicates for all the objects in the repository, and only the ones related to the objects appearing in the current scene is evaluated. For SORNet, we give it the canonical patches corresponding to the objects appearing in the scene.

We report the average F-1 score on Fig.~\ref{fig:quant}. We can see that in the frozen embedding scenario, the baselines' performance is significantly worse than SORNet. This indicates existing pre-training methods do not work in a plug-and-play fashion on spatial reasoning tasks. The learned representations simply do not contain information necessary for predicting spatial relationships among objects. In the finetuned embedding case, the baselines are performing much better, but SORNet still outperforms all baselines, especially on the Kitchen data which contains more complex objects and scenes. This demonstrates the advantage of an object-centric network.

\begin{figure}
    \centering
    \includegraphics[width=0.9\linewidth]{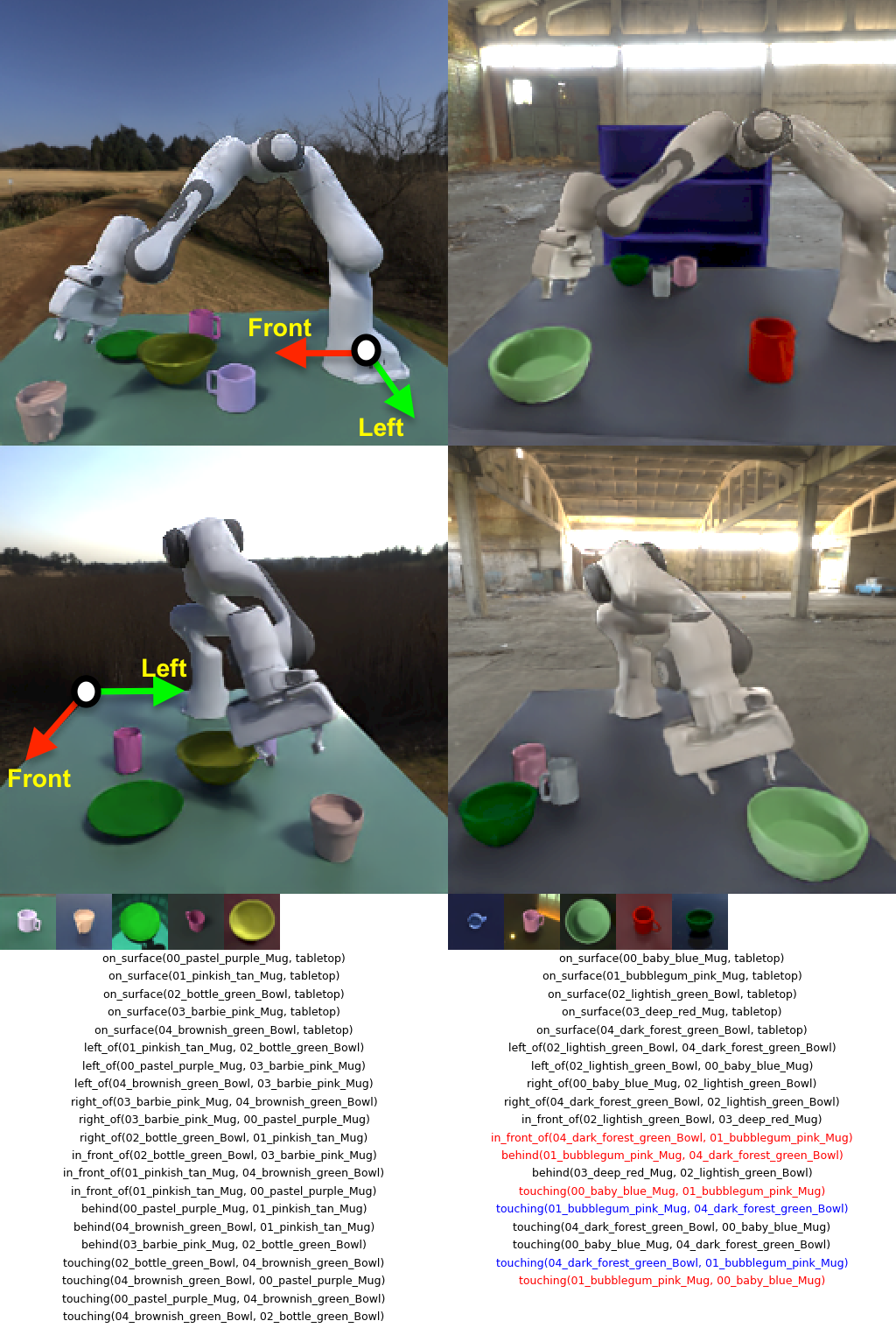}
    \caption{Qualitative predicate classification results with SORNet trained on the simulated kitchen dataset and tested on held-out objects. Each column is a different scenario. The first and second row shows the side and front view of the scene respectively, followed by the canonical views of 5 query objects. Black text denotes correctly labeled true predicates; blue text denotes \textcolor{blue}{false positive predictions}; and red text denotes \textcolor{red}{false negatives}. True negatives are not shown due to limited space.}
    \label{fig:qual}
\end{figure}

Further, the design of SORNet allows it to be applied zero-shot, i.e. without any additional annotation, to objects that are unseen during training. This is reflected by the \textbf{SORNet (0-shot)} model in Fig.~\ref{fig:quant}. This cannot be achieved by any of the baselines since their representations is not object-specific, so they would require additional training/fine-tuning in order to work on novel objects. Fig.~\ref{fig:qual} shows some qualitative results of the zero-shot model on the kitchen data. Note that both the color and the shape (category) of all objects do not appear in the training set.

\begin{figure}
    \centering
    \includegraphics[width=0.49\linewidth]{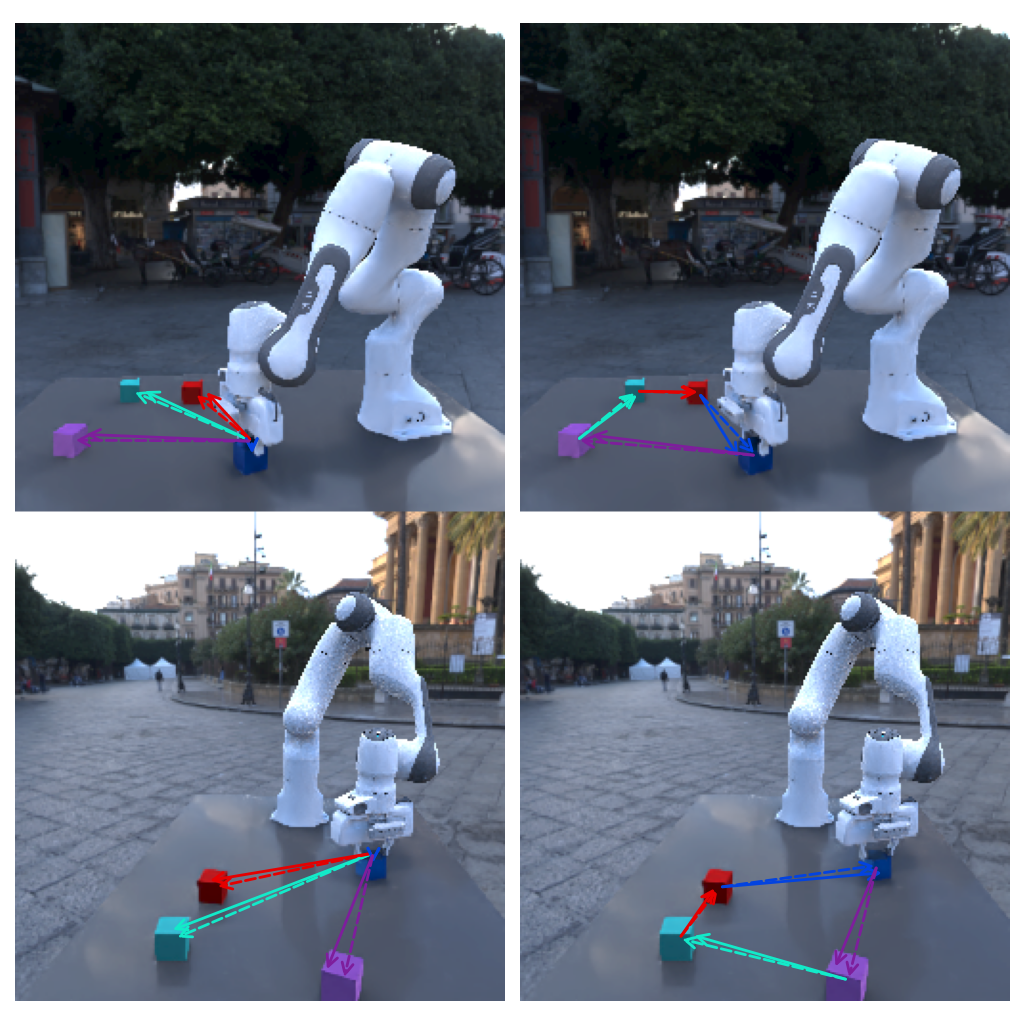}
    \includegraphics[width=0.49\linewidth]{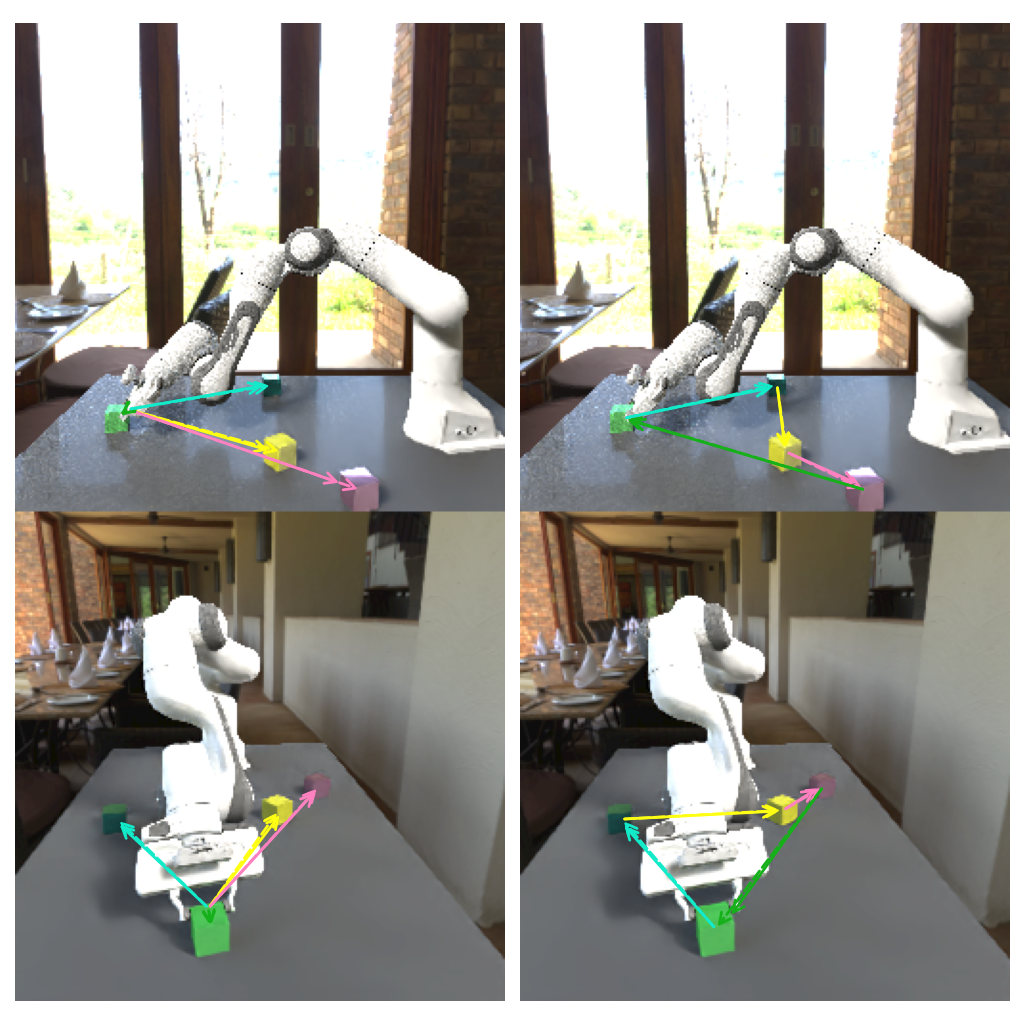}
    \includegraphics[width=0.49\linewidth]{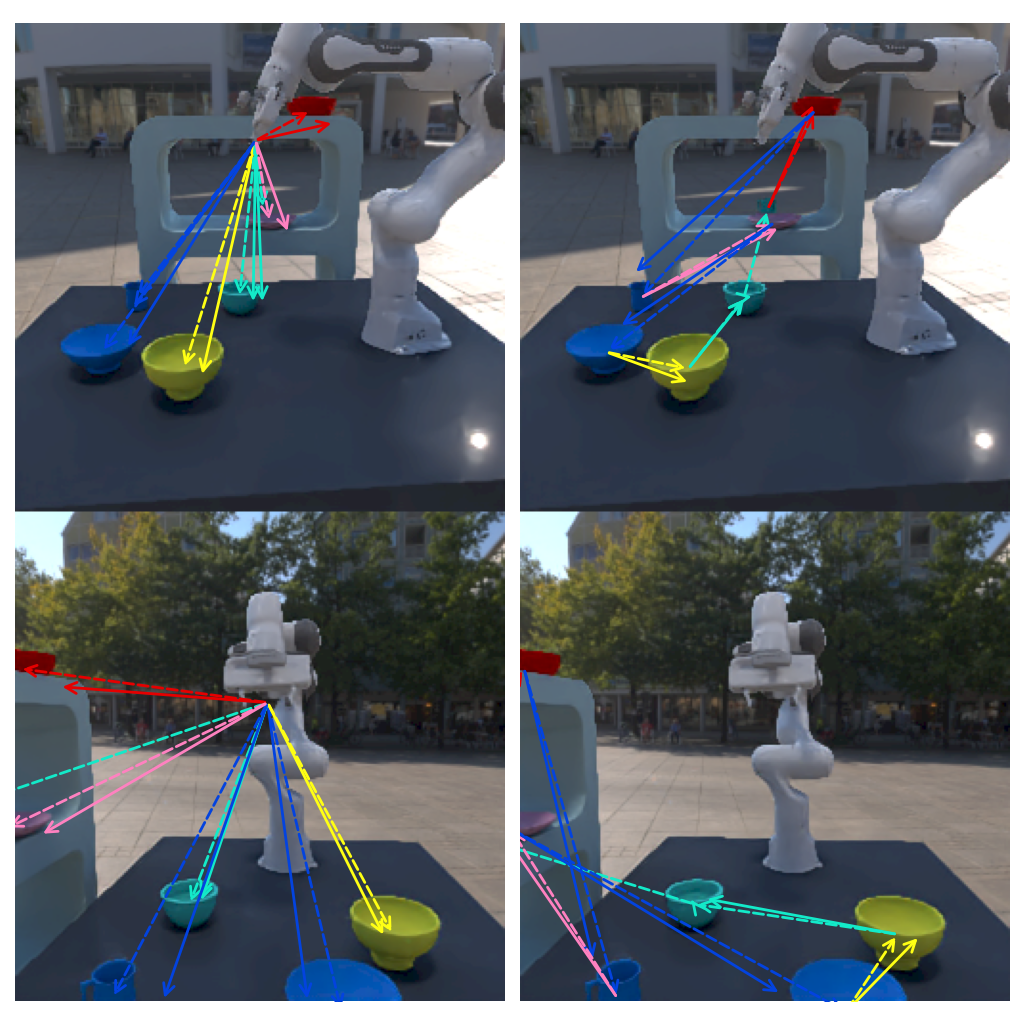}
    \includegraphics[width=0.49\linewidth]{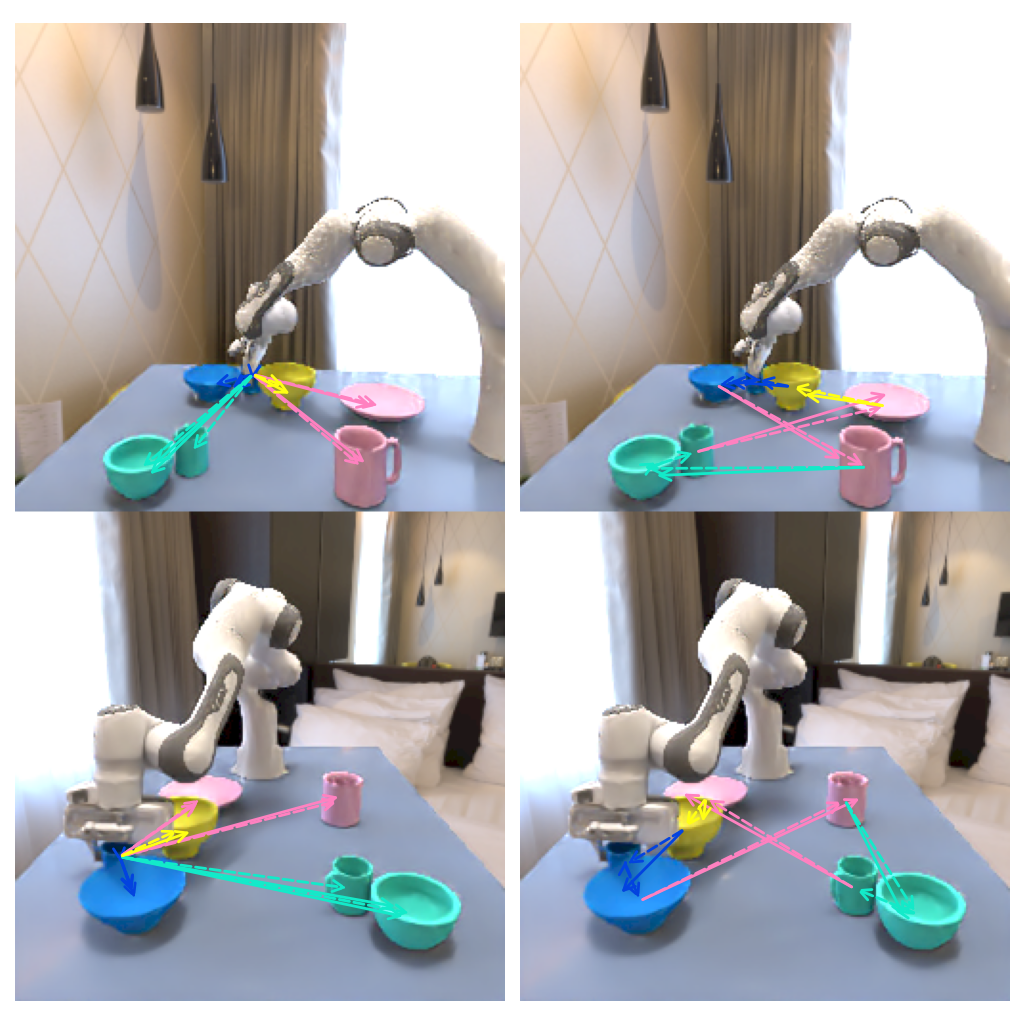}
    \caption{Relative Direction Prediction using pretrained SORNet embeddings. The color of the arrow corresponds to the color of the target. Solid arrows are the predictions and dashed arrows are the ground truths. Not all predictions are visualized in order to keep the plot clean. The first and third columns show the relative direction from the robot's end effector to the object centers. The second and fourth columns show the relative direction between object centers.}
    \label{fig:dirreg}
\end{figure}

\subsection{Relative Direction Regression} \label{sec:reg}
In additional to predicting logical relations, the object embeddings learned by SORNet can also be used to predict continuous spatial information.
To demonstrate this, we use apply SORNet to predict the relative 3D direction between entities. Specifically, we trained a regressor (same architecture as the classifier in Sec.~\ref{sec:classifier}) to predict the continuous vector between two object centers (Obj-Obj) or the direction the end effector should move to reach a certain object (EE-Obj). The $x,y,z$ components of the continuous vector are treated as ``predicates" with continuous values and trained with L2 loss. The regressor predicts both the directions from the robot's end-effector to the object centers (EE-Obj), and the directions between the object centers (Obj-Obj).

Results are summarized in Fig~\ref{fig:quant}. We compare to the same baselines and employ the same two evaluation settings as in Sec.~\ref{sec:predicate}. SORNet is able to outperform competing pretraining methods (CLIP, MVP, R3M and MAE) in a similar fashion. This demonstrates that SORNet's representation transfers much better to manipulation scenarios where more precise spatial information is crucial. We can also apply SORNet in a zero-shot fashion to scenes with novel objects. Fig.~\ref{fig:dirreg} shows some qualitative prediction of the zero-shot SORNet model.

\begin{table}
    \centering
    \footnotesize
    \begin{tabular}{cccc}
        \toprule
         & MDETR \cite{kamath2021mdetr} & MDETR-oracle \cite{kamath2021mdetr} & \textbf{sornet{}(ours)} \\
        \midrule
        ValA Accuracy & 84.950 & 97.944 & \textbf{99.006} \\
        ValB Accuracy & 59.627 & 98.052 & \textbf{98.222} \\
        \bottomrule
    \end{tabular}
    \vspace{2mm}
    \caption{Zero-shot relation classification accuracy on CLEVR-CoGenT \cite{johnson2017clevr}. The MDETR-oracle model has seen all the objects during training, where as MDETR and SORNet have only see objects in condition A. SORNet takes canonical views as queries whereas MDETR takes text queries.}
    \label{tab:clevr}
\end{table}

\subsection{Compositional Generalization on CLEVR-CoGenT}
We also evaluate our approach on a variant of the CLEVR dataset~\cite{johnson2017clevr}, a well established benchmark for visual reasoning. CLEVR contains rendered RGB images with at most 10 objects per image. There are 96 different objects in total (2 sizes, 8 colors, 2 materials, 3 shapes). Each image is labeled with 4 types of spatial relations (right, front, left, behind) for each pair of objects.

Specifically, we use the CoGenT version of the dataset, which stands for Compositional Generalization Test, where the data is generated in two different conditions. In condition A, cubes are gray, blue, brown, or yellow and cylinders are red, green, purple, or cyan. Condition B is the opposite: cubes are red, green, purple, or cyan and cylinders are gray, blue, brown, or yellow. Spheres can be any color in both conditions. The models are trained on condition A and evaluated on condition B. The training set (trainA) contains 70K images and the evaluation set (valB) contains 15K images. Several prior works \cite{yi2018neural,kamath2021mdetr} show significant generalization gap on CLEVR-CoGenT caused by the visual model learning strong spurious biases between shape and color.

\begin{figure}
    \centering
    \includegraphics[width=\linewidth]{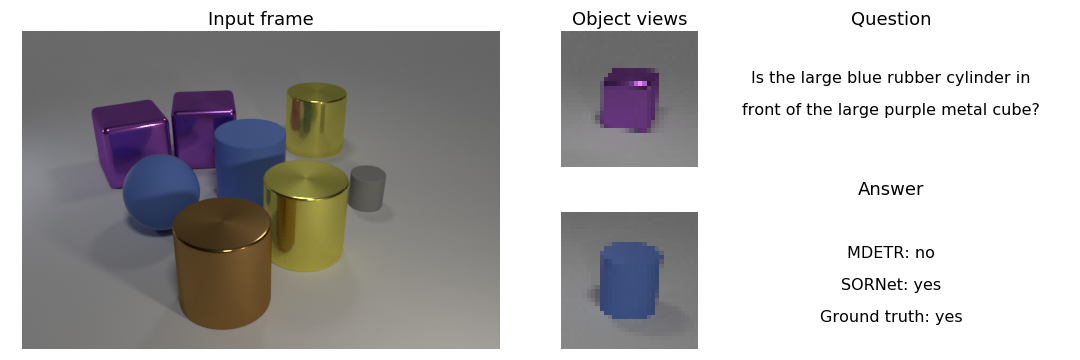}
    \includegraphics[width=\linewidth]{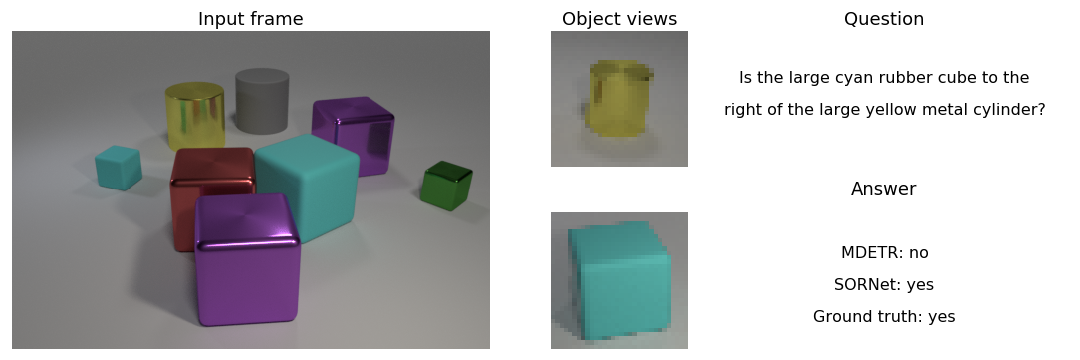}
    \includegraphics[width=\linewidth]{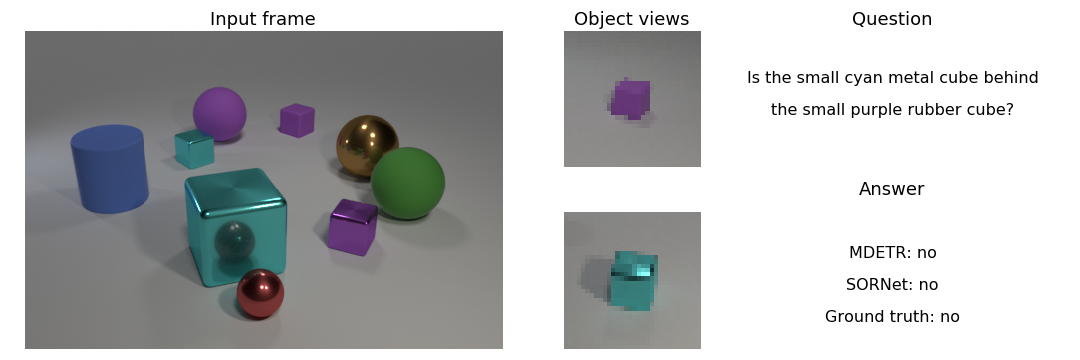}
    \includegraphics[width=\linewidth]{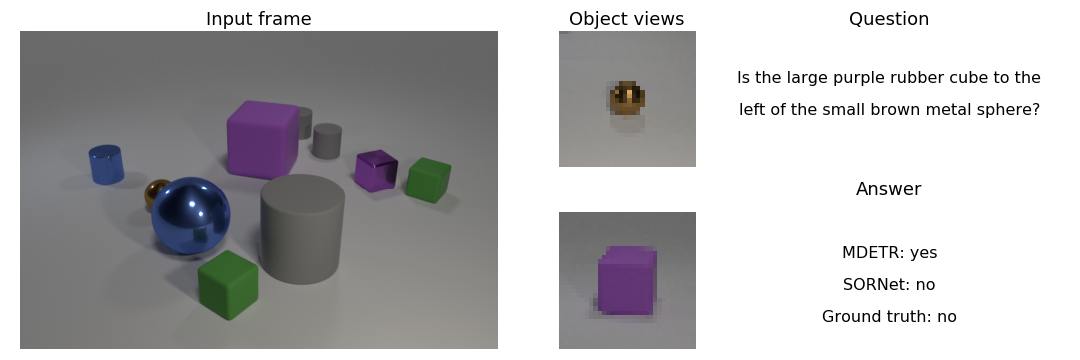}
    \caption{Relation classification in CLEVR-CoGenT. In addition to color, SORNet is able to distinguish objects based on shape, size and material. It is also able to deal with heavy occlusion. Note that the object patches provided to SORNet can have very different appearance than the corresponding objects in the input frame.}
    \label{fig:clevr}
\end{figure}

We generate a question for each spatial relation in the image, e.g. ``\textit{Is the large red rubber cube in front of the small blue metal sphere?}'' We filter out any query that is ambiguous, e.g. if there were two large red cubes, one in front and one behind the small blue sphere. This results in around 2 million questions for both valA and valB sets. We compare against MDETR \cite{kamath2021mdetr}, which reports state-of-the-art zero-shot result on CLEVR-CoGenT, i.e. there is no fine-tuning on any example from condition B. The results are summarized in Table~\ref{tab:clevr}. Our model performs drastically better on classifying spatial relations of unseen objects and shows a much smaller generalization gap between valA and valB sets.

Unlike MDETR which takes text queries, our model takes visual queries in the form of canonical object views (i.e. 2 canonical views for the objects mentioned in the question). To eliminate the influence of those factors, we report the performance of the MDETR model trained on the full CLEVR dataset, denoted as MDETR-oracle. We can see that although SORNet is only trained on condition A, it is able to achieve similar performance to MDETR-oracle. The zero-shot generalization ability of our model can potentially be combined with other reasoning pipelines to improve generalization performance on other types of queries as well.

Fig.~\ref{fig:clevr} shows examples of spatial relation classification on CLEVR-CoGenT. These examples demonstrate that SORNet is able to identify objects not only using color cues, but also shape (e.g. blue sphere vs blue cylinder in the topmost example), size (e.g. small cyan cube vs big cyan cube in the second from top example) and material (e.g. small purple metal cube vs small purple rubber cube in the third from top example).
We also visualize the relevant canonical object views provided to the model. These can have a very different appearance from the corresponding objects in the input image. It is more appropriate to consider these canonical views as a visual replacement for natural language, rather than the result of object detection or segmentation.

%% file: 07casestudy.tex
\section{Applications}
Finally, we show a set of experiments that demonstrate the application of SORNet in various simulated and real-world manipulation scenarios.

\begin{figure}
    \includegraphics[width=\linewidth]{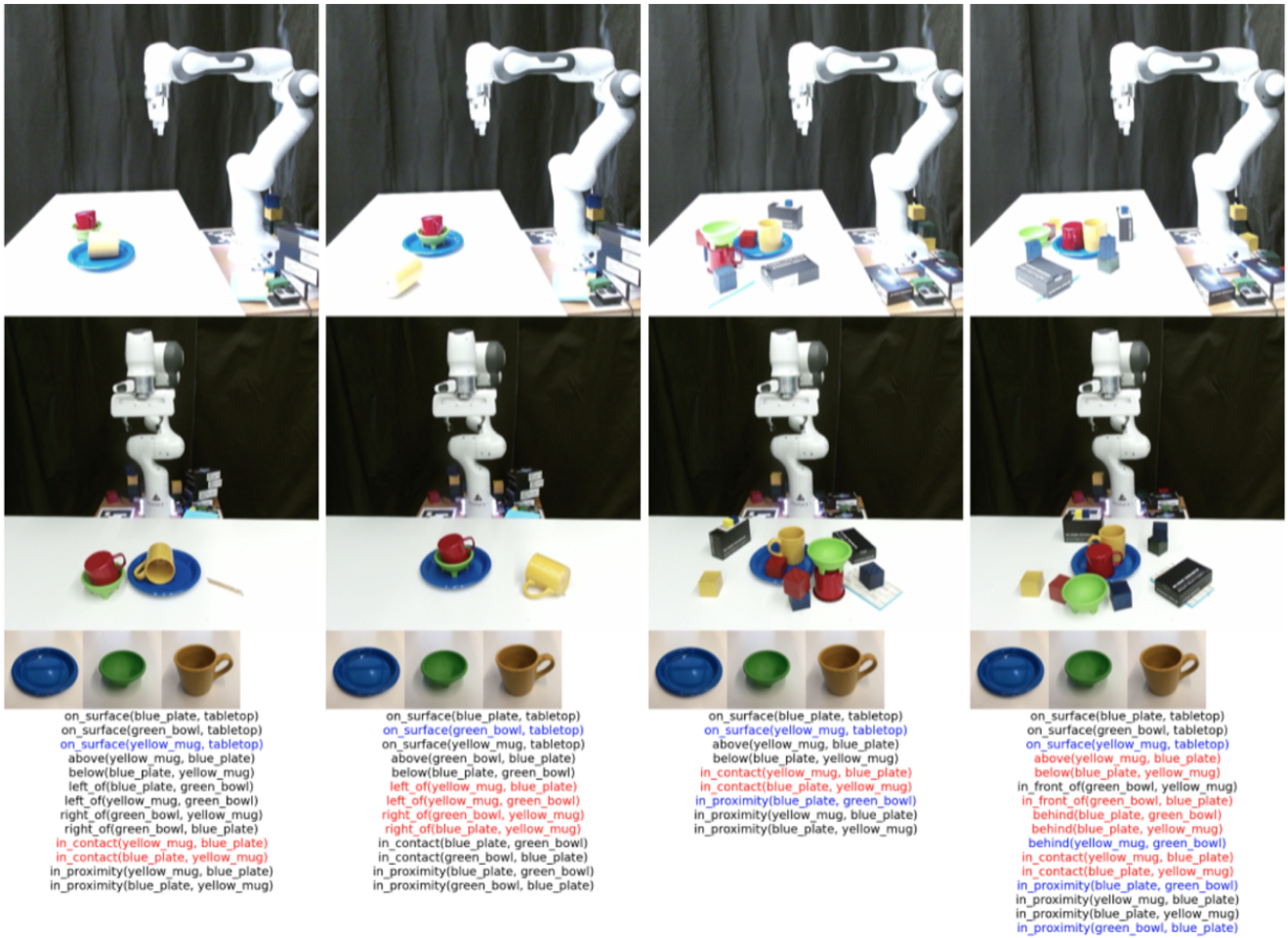}
    \caption{Predicate classification by SORNet trained on the simulated kitchen dataset and tested on held-out real-world objects. Each column is a different scenario. The first and second row shows the side and front view of the scene respectively, followed by the canonical views of the query objects (``blue\_plate and ``red\_mug"). Black text denotes correctly labeled true predicates; blue text denotes \textcolor{blue}{false positive predictions}; and red text denotes \textcolor{red}{false negatives}. We can see that SORNet produces accurate labels in many scenes, and shows strong transfer even to scenes and objects that are very out of its training distribution.}
    \label{fig:sornet-preds-real-world}
\end{figure}

\subsection{Sim-to-real Transfer}

SORNet embeddings can transfer from simulation to the real world with no fine tuning. In order to demonstrate this, we performed a set of experiments where we captured a range of scenes containing novel, unseen objects in previously unseen colors, from our held-out test objects set. In each scene, we picked one random pair of objects and compared manually-labeled predicates to those predicted by SORNet.

Fig.~\ref{fig:sornet-preds-real-world} shows some examples of scenes from the real-world testing dataset. Importantly, these objects were randomly chosen from the real-world. Two cameras are placed without any extrinsic calibration to view the scene from side and front views. SORNet with 2 views and trained only with RGB is used for this experiment. The canonical views are pictures taken using the ``square" mode on a smart phone. Each column shows a particular state of the world and the predicate classification by SORNet. To simplify the illustration we chose to show the predicates of only two objects ``red\_mug" and ``blue\_plate." Black colored text denotes true positive predictions, blue colored text denotes false positive predictions and red text denotes false negatives. SORNet was never trained on real-world observations to perform these predictions. We can see that SORNet produces accurate classification in many scenes, and shows strong transfer even to scenes and objects that are out of its training distribution.

\subsection{Open Loop Planning}

In this demo, we incorporate SORNet as a part of an open-loop planning pipeline in a real-world manipulation scenario. Specifically, given an initial frame, we use the predicates predicted by SORNet to populate a state vector. A task and motion planner takes the state vector and desired goal (formulated as a list of predicate values to be satisfied), and outputs a sequence of primitive skills. The robot then executes this sequence of skills in an open loop fashion. This demonstrate that how SORNet can be applied to sequential manipulation of unseen object in a zero-shot fashion, i.e. without any fine-tuning on the test objects. Please refer to our website for the demo video.

\subsection{Visual Servoing}

We can use SORNet to predict a 3D direction which guides the robot's end effector towards a specified object. Specifically, we train a regressor (a 2-layer MLP) on top of SORNet's pretrained embeddings to predict the direction (a 3D unit vector) and distance (a 1D scalar) from the robot's end effector to each object in the scene (see Sec.~\ref{sec:reg} and Fig.~\ref{fig:dirreg} for examples). In the demo, the robot's end effector moves towards a target object using position control. The direction and distance for the movement is predicted by the regressor in an online fashion. The SORNet embedding used as input for the regressor is also produced online. This experiment demonstrates that even though our model is not trained on any metric task, the learned embeddings contain continuous 3D spatial information accurate enough to guide the robot's motion for primitive skills, e.g. approach an object. Please refer to our website for the demo video.

%% file: 09conclusion.tex
\section{Conclusion}
We proposed \textbf{SORNet} (\textbf{S}patial \textbf{O}bject-Centric \textbf{R}epresentation \textbf{Net}work) that learns object-centric representations from RGB images. We show that the object embeddings produced by SORNet capture spatial relations which can be used in a downstream tasks such as spatial relation classification, skill precondition classification and relative direction regression. Our method works on scenes with an arbitrary number of unseen objects in a zero-shot fashion. With real-world robot experiments, we demonstrate how SORNet can be used in manipulation of novel objects.

%% file: 10supplement.tex
\section*{Supplementary Material}
\beginsupplement

\input{08analysis}

%------------------------------------------------------------------------
\section{Model Architecture and Training}
\subsection{SORNet Architecture}
The embedding network consists of 6 attention layers. Each layer is an 8-head self attention block. The width of the block, i.e. the dimension for the token vectors, is set to 512. Each block has an self attention layer followed by a 2-layer MLP whose hidden dimension is 4 times the width (2048). Layer norm is applied to the inputs to the attention and the MLP. Residual connection is applied after the attention and the MLP. We use the GELU activation function for the MLP. Note that this is a smaller transformer compared to the ViT-B model in \cite{dosovitskiy2020image}. We found that this lightweight model trains faster and is less prone to overfitting.

The size of the image patches in the input sequence is $16\times 16$. The size of the canonical object views is $64\times 64$, so each object will correspond to 16 tokens. We find it is important to use smaller patch size and bigger canonical object views, especially for the Kitchen data. Although this means longer input sequence length and more memory usage, the increased spatial resolution is crucial to deal with the varying object size in the Kitchen data. We compensate for the additional memory usage by using a shallower (6-layer) transformer encoder as opposed to the 12-layer one in \cite{dosovitskiy2020image}.

The readout networks are 2-layer MLPs whose hidden layer dimension is 512. It also uses the GELU activation function.

\subsection{Training Settings}
All models are trained using the SGD optimizer (momentum set to 0.9) on 4 GPUs with 32G memory. The batch size on a single GPU is 64, so the effective batch size is $64\times 4=256$. The learning rate is set to 0.0001. The models are trained for 40 epochs on the Leonardo data and 80 epochs on the Kitchen data. The predicate classification models are trained with binary cross entropy loss. The direction regression models are trained with L1 loss.

\subsection{Baseline Architecture}
All baselines are trained to predict predicates/directions for all test objects, whether they appear in the current scene or not. For example, in the Kitchen data there are 14 test objects, so the readout MLPs for the EE-Obj direction regression task will predict $3\times 14=42$ outputs (for $x,y,z$ components of the directions). We only evaluate on the objects that appears in the scene and ignore all other outputs.

\paragraph{CLIP}
We use the ViT-B/16 model provided by the official CLIP implementation at \url{https://github.com/openai/CLIP}. The model is pretrained on a large, non-public image-text alignment dataset. The model has 12 layers of 12-head self-attention and a width of 768. The training hyperparameters are the same as SORNet.

\paragraph{R3M}
We use the model with ResNet50 backbone from the implementation by R3M's authors \url{https://github.com/facebookresearch/r3m}. The model is pretrained on the Ego4D dataset using the contrastive objective proposed by \cite{nair2022r3m}. ResNet50 is a pretty standard backbone whose architecture can be found in \cite{he2016deep}. We use a larger learning rate of 0.001 for R3M due to the different backbone architecture.

\paragraph{MVP}
We use the ViT-S model provided by the implementation by MVP's authors \url{https://github.com/ir413/mvp}. The model is trained on a combined dataset referred to as HOI in \cite{xiao2022masked}. The model architecture is the same as the ViT-B/32 model in \cite{dosovitskiy2020image} except that the width of the model is halved to 384. We use the same training hyperparameters as SORNet.

\paragraph{MAE}
We use the ViT-B/16 model in the official MAE implementation \url{https://github.com/facebookresearch/mae}. It has the same architecture as the CLIP model we used. The model is pre-trained on ImageNet-1K with masked recontruction objective. We also provide a variant of the MAE model finetuned on our data, denoted as \textbf{MAE-sor}. This model is trained to reconstruct images from the Leonardo or Kitchen data. Interestingly, we did not find a big difference compared to the MAE model pre-trained on ImageNet, which indicates that dataset bias is not the key issue in our comparisons.

%------------------------------------------------------------------------
\section{Dataset Details}
\subsection{List of Object Categories}
Table~\ref{tab:categories} lists the object categories in the Kitchen dataset along with the number of objects in each category. Heldout object categories are highlighted in red.

\subsection{List of Predicates}
Table~\ref{tab:predicates} shows the list of logical predicate types included in the Leonardo and Kitchen dataset respectively. $\cdot$ denote positions for object arguments. Unary relations have a single $\cdot$ while binary relations have two $\cdot$s

\begin{table}
    \centering
    \small
    \begin{tabular}{|l|l|l|l|l|}
    \toprule
        Basket (7) & BeerBottle (9) & Book (12) & Bottle (12) & \textcolor{red}{Bowl (14)} \\
        Calculator (12) & Candle (12) & CellPhone (9) & CerealBox (14) & ComputerMouse (7) \\
        Controller (2) & Cup (24) & Donut (5) & Fork (16) & Fruit (2) \\
        Hammer (9) & Knife (8) & Marker (1) & MilkCarton (11) & \textcolor{red}{Mug (26)} \\
        Pan (20) & Pen (4) & PillBottle (6) & Plate (12) & PowerStrip (4) \\
        Scissors (2) & SoapBottle (3) & SodaCan (14) & Spoon (10) & Stapler (16) \\
        Teapot (12) & VideoGameController (4) & WineBottle (11) & & \\
    \bottomrule
    \end{tabular}
    \vspace{2mm}
    \caption{List of 33 object categories in the Kitchen dataset. The number of objects in each category is included in the parenthesis. The heldout (test) object categories are labeled in red.}
    \label{tab:categories}
\end{table}

\begin{table}
\begin{subfigure}{0.3\linewidth}
\centering
\includegraphics[width=\linewidth]{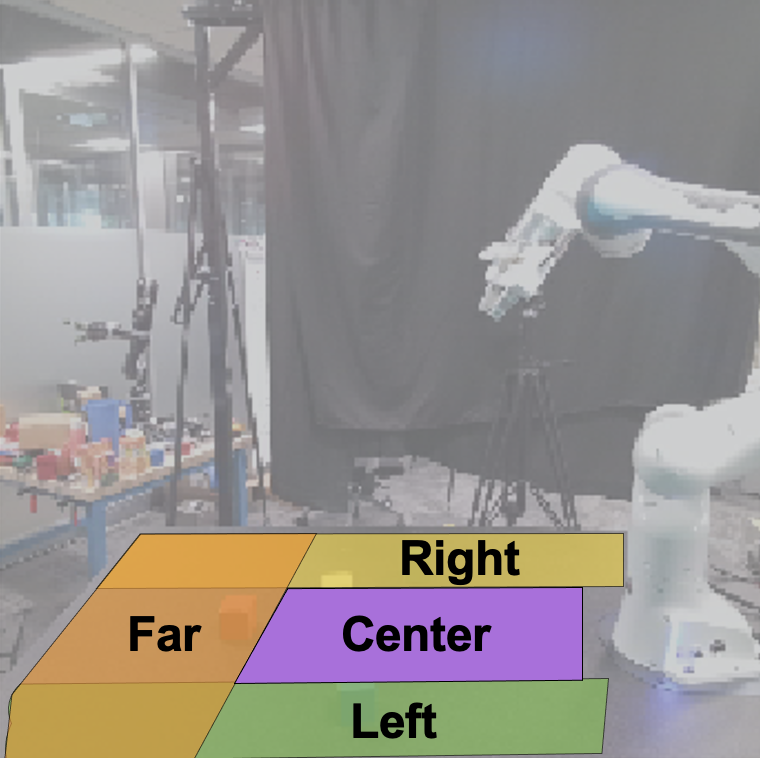}
\caption{Table regions}
\end{subfigure}
\hfill
\begin{subfigure}{0.34\linewidth}
\centering
\small
\begin{tabular}{|l|}
\toprule
on\_surface($\cdot$, left) \\
on\_surface($\cdot$, right) \\
on\_surface($\cdot$, far) \\
on\_surface($\cdot$, center) \\
has\_obj(robot, $\cdot$) \\
top\_is\_clear($\cdot$) \\
in\_approach\_region(robot, $\cdot$) \\
\midrule
stacked($\cdot$, $\cdot$) \\
aligned\_with($\cdot$, $\cdot$) \\
\bottomrule
\end{tabular}
\vspace{2mm}
\caption{Predicates in Leonardo data}
\end{subfigure}
\begin{subfigure}{0.34\linewidth}
\centering
\small
\begin{tabular}{|l|}
\toprule
on\_surface($\cdot$, tabletop) \\
on\_surface($\cdot$, shelf) \\
\midrule
abvoe($\cdot$, $\cdot$) \\
below($\cdot$, $\cdot$) \\
left\_of($\cdot$, $\cdot$) \\
right\_of($\cdot$, $\cdot$) \\
in\_front\_of($\cdot$, $\cdot$) \\
behind($\cdot$, $\cdot$) \\
in\_proximity($\cdot$, $\cdot$) \\
in\_contact($\cdot$, $\cdot$) \\
\bottomrule
\end{tabular}
\vspace{2mm}
\caption{Predicates in Kitchen data}
\end{subfigure}
\caption{(a) a table divided into 4 regions (left, right, far and center) with respect to the robot. (b) List of predicate types in the Leonardo dataset. (c) List of predicate types in the Kitchen dataset. $\cdot$ denote positions for object arguments.}
\label{tab:predicates}
\end{table}

%% file: 08analysis.tex
\section{Analysis}
We performed some extra experiments to show what exactly SORNet is learning and what it is attending to.

\subsection{Multiview and Depth Inputs}
\begin{figure}
    \centering
    \includegraphics[width=0.4\linewidth]{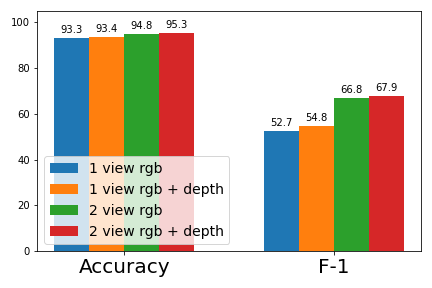}
    \includegraphics[width=0.59\linewidth]{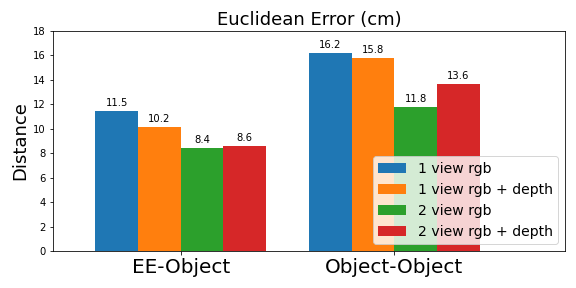}
    \caption{Predicate classification and direction regression results on the kitchen dataset with multiview and depth inputs. SORNet is able to leverage additional data to improve its performance.}
    \label{fig:cls_inputs}
\end{figure}

Fig.~\ref{fig:cls_inputs} shows the improvements SORNet can get from using multiple viewpoints and depth. The transformer backbone enables SORNet to incorporate additional sensor inputs without any architectural changes.

\subsection{Different Number of Objects}
\begin{figure}
    \centering
    \includegraphics[width=0.4\linewidth]{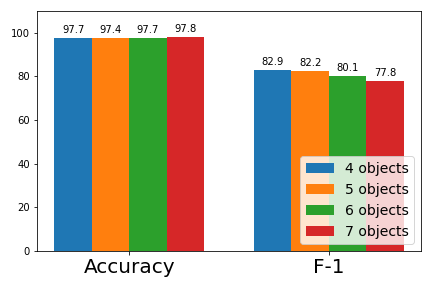}
    \caption{SORNet generalizes to scenes with different number of objects with almost no performance drop.
    Results are shown for predicate prediction and for overall F1 score.}
    \label{fig:nobj}
\end{figure}

We tested SORNet, which was only trained on 4-object scenes, to scenes with 5 to 7 objects. We can see from Fig.~\ref{fig:nobj} that SORNet generalizes to scenes with different number of objects with almost no performance drop.

\begin{figure}
    \centering
    \includegraphics[width=\linewidth]{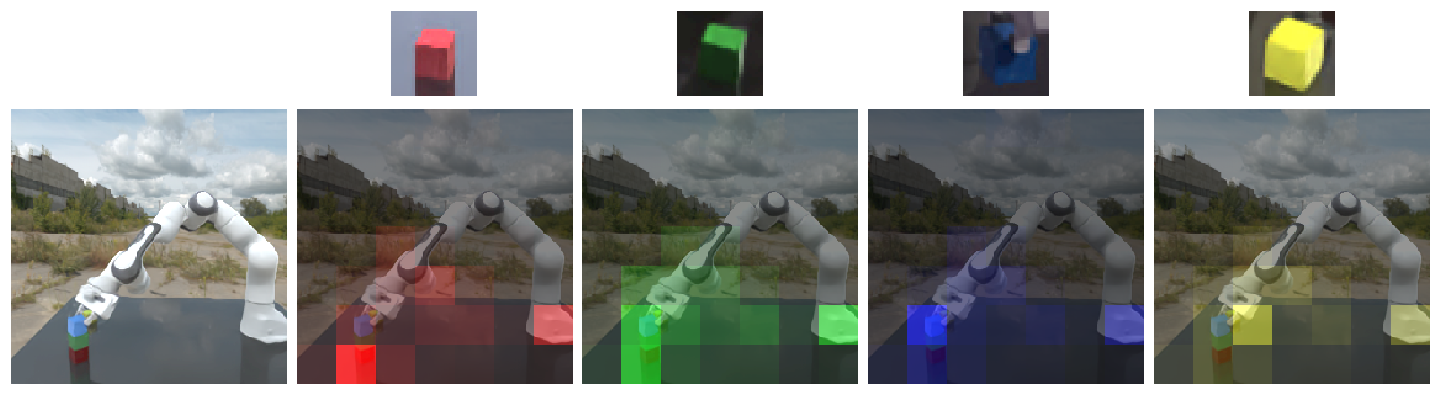}
    \includegraphics[width=\linewidth]{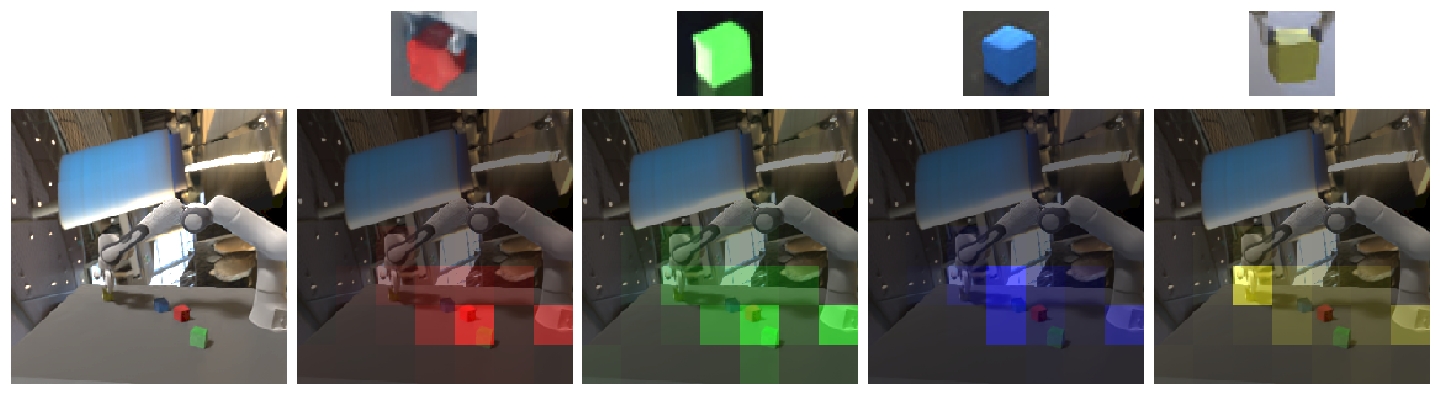}
    \includegraphics[width=\linewidth]{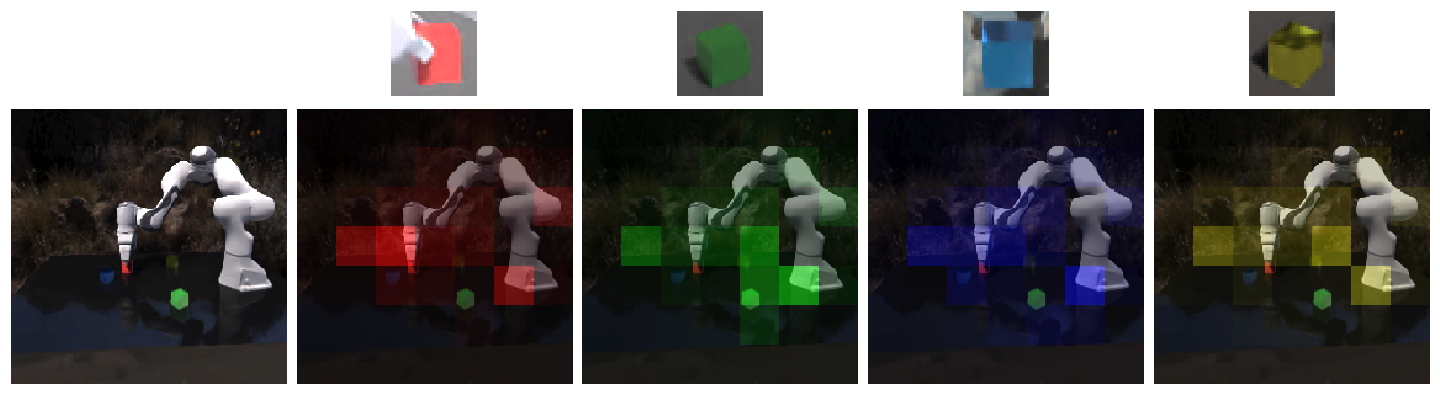}
    \includegraphics[width=\linewidth]{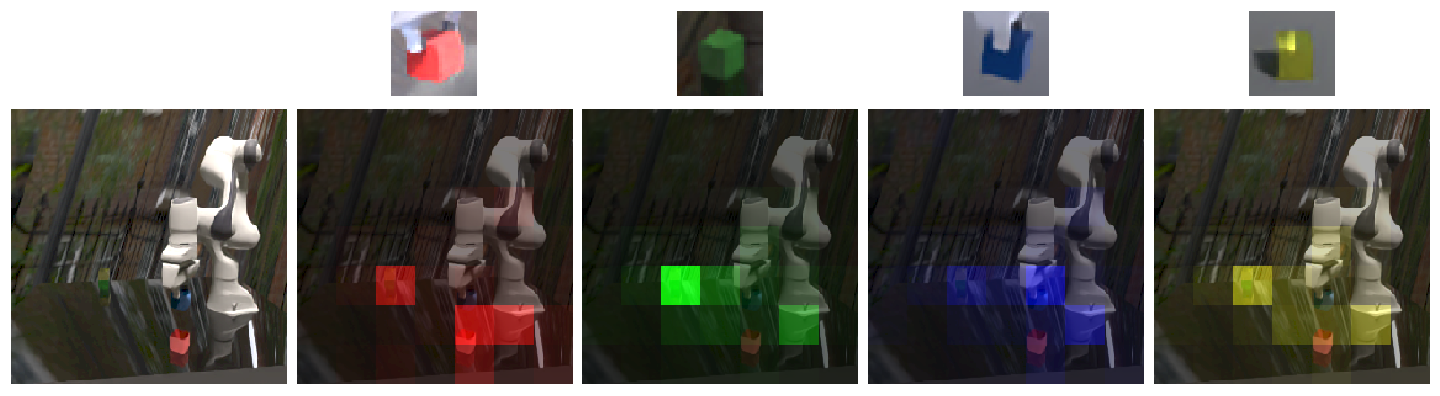}
    \includegraphics[width=\linewidth]{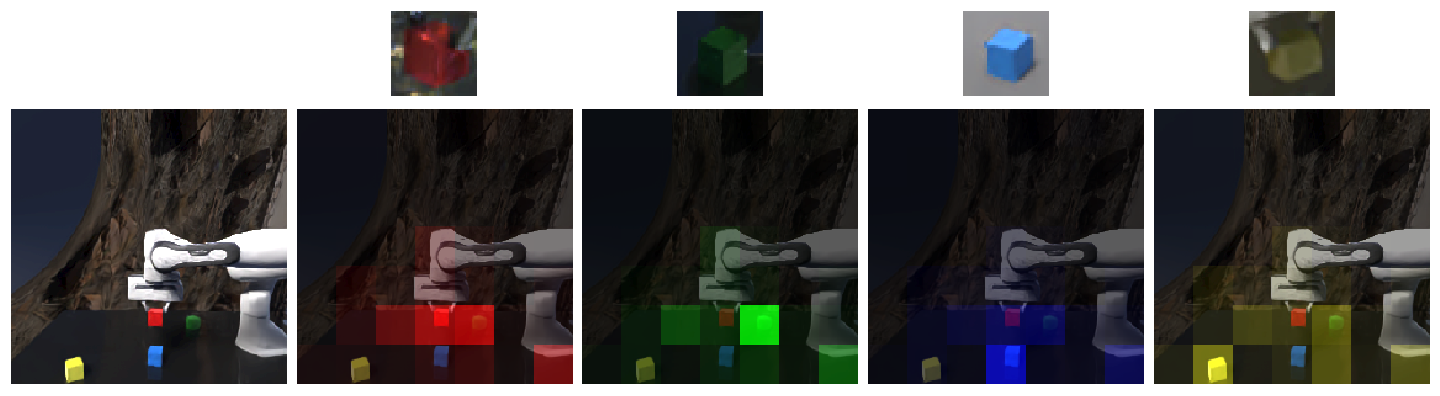}
    \caption{Visualization of the attention learnt by SORNet. Leftmost column is the input frame and remaining columns visualize the attention weights of the object tokens over the context patches. The corresponding canonical object view is shown on top of each attention visualization. We can see that SORNet learns to attend to not only the object of interest but also the robot and other objects as well, while ignoring irrelevant background.}
    \label{fig:attn}
\end{figure}

\subsection{Attention Visualization} \label{sec:attn}
Fig.~\ref{fig:attn} visualizes the attention weights learned by the visual transformer model in order to obtain the object-centric embeddings. More specifically, we take the normalized attention weights from the tokens corresponding to the canonical object views to the tokens corresponding to the context patches and convert the weights into a colormap, where the intensity of the colormap corresponds to the magnitude of the attention weight over that patch. We then overlay the colormap with the input image. We can see that while the model puts the highest attention to the patch containing the object of interest, it also learns to pay attention to the robot arm and other objects, while ignoring irrelevant background. We also visualize the canonical object patches given to the model, which can look wildly different from the same objects in the image. The model needs to associate the canonical view with the input view of the object under different lighting conditions and occlusions.